\documentclass[sigconf]{aamas}  

\AtBeginDocument{%
  \providecommand\BibTeX{{%
    \normalfont B\kern-0.5em{\scshape i\kern-0.25em b}\kern-0.8em\TeX}}}
    
\usepackage{flushend} 
\usepackage{microtype}
\usepackage{booktabs}  
\usepackage[]{algpseudocode}
\usepackage{amsmath}
\usepackage{color}
\usepackage{soul}
\usepackage{xspace}
\usepackage{cleveref}
\usepackage{dsfont}
\usepackage{bbm}
\usepackage{nicefrac}  
\usepackage[scaled=.8]{beramono}  
\usepackage{amsthm}
\usepackage{amsfonts}
\usepackage{amssymb}
\usepackage{array}
\usepackage{mathtools}
\usepackage{caption}
\usepackage{subcaption}
\usepackage{fontawesome}
\usepackage{thmtools}
\usepackage{thm-restate}
\usepackage{xspace}
\usepackage{lipsum}
\usepackage{pifont}
\usepackage{makecell}
\usepackage{tablefootnote}
\usepackage[tight-spacing=true]{siunitx}
\usepackage{multirow}
\usepackage{afterpage}
\usepackage{tabularx} 
\usepackage{url}
\usepackage{tikz}
\usepackage{gensymb}  
\usepackage{algorithm}

\Crefname{figure}{Figure}{Figures}
\crefname{figure}{Figure}{Figures}
\crefname{table}{Table}{Tables}
\crefformat{table}{Table~#1}
\crefformat{equation}{(#1)}
\Crefformat{equation}{(#1)}
\crefformat{algorithm}{Algorithm~#1}
\crefformat{appendix}{Appendix~#1}
\Crefformat{appendix}{Appendix~#1}
\crefformat{section}{section~#1}
\Crefformat{section}{Section~#1}

\DeclareMathOperator*{\E}{\mathbb{E}}

\newcommand{\header}[1]{\textbf{#1}\,\,}
\newcommand{\worker}{\mathcal{W}} 
\newcommand{\manager}{\mathcal{M}} 
\newcommand{\meta}[1]{\widetilde{#1}}
\newcommand{\Ptask}{$\mathcal{P}_{\text{task}}$\xspace}
\newcommand{\Padvice}{$\meta{\mathcal{P}}_{\text{advice}}$\xspace}

\newcommand\sdots{\hbox to 1em{.\hss.\hss.}} 

\newcommand{\cmark}{\ding{51}} 
\newcommand{\xmark}{\ding{55}} 

\usetikzlibrary{decorations.pathreplacing,calc}
\newcommand{\tikzmark}[1]{\tikz[overlay,remember picture] \node (#1) {};}
\newcommand*{\AddNote}[4]{%
    \begin{tikzpicture}[overlay, remember picture]
        \draw [decoration={brace,mirror,amplitude=0.5em},decorate,thick,black]
            ($(#3)!(#1.north)!($(#3)-(0,1)$)$) --  
            ($(#3)!(#2.south)!($(#3)-(0,1)$)$)
                node[align=center,text width=2.5cm,pos=0.5,anchor=east,xshift=0.85cm]{#4};
    \end{tikzpicture}
}

\newif\ifcomments
\commentsfalse 

\ifcomments
	\newcommand{\dXX}[1]{\color{red}DK: (#1)\color{black}\xspace}  
	\newcommand{\soXX}[1]{\color{olive}SO: (#1)\color{black}\xspace}  
	\newcommand{\XX}[1]{\color{orange}JH: (#1)\color{black}\xspace}  
	\newcommand{\mXX}[1]{\color{cyan}ML: (#1)\color{black}\xspace}  
	\newcommand{\gtXX}[1]{\color{purple}GT: (#1)\color{black}\xspace}  
	\newcommand{\mrXX}[1]{\color{blue}MR: (#1)\color{black}}  
	\newcommand{\slXX}[1]{\color{magenta}SL: (#1)\color{black}}  
	\newcommand{\ghXX}[1]{\color{olive}GH: (#1)\color{black}}  
	\newcommand{\smXX}[1]{\color{red}SM: (#1)\color{black}}  
	\newcommand{\mcXX}[1]{\color{olive}MC: (#1)\color{black}}  
\else
    \newcommand{\dXX}[1]{}  
	\newcommand{\soXX}[1]{}  
	\newcommand{\XX}[1]{}  
	\newcommand{\mXX}[1]{}  
	\newcommand{\gtXX}[1]{}  
	\newcommand{\mrXX}[1]{}  
	\newcommand{\slXX}[1]{}  
	\newcommand{\ghXX}[1]{}  
	\newcommand{\smXX}[1]{}  
	\newcommand{\mcXX}[1]{}  
\fi

\renewrobustcmd{\bfseries}{\fontseries{b}\selectfont}
\renewrobustcmd{\boldmath}{}
\newrobustcmd{\B}{\bfseries}
\DeclareMathVersion{normal2}

\setcopyright{ifaamas}  
\copyrightyear{2020} 
\acmYear{2020} 
\acmDOI{} 
\acmPrice{} 
\acmISBN{} 
\acmConference[AAMAS'20]{Proc.\@ of the 19th International Conference on Autonomous Agents and Multiagent Systems (AAMAS 2020)}{May 9--13, 2020}{Auckland, New Zealand}{B.~An, N.~Yorke-Smith, A.~El~Fallah~Seghrouchni, G.~Sukthankar (eds.)}  

\begin{document}
   
\title{Learning Hierarchical Teaching Policies for Cooperative Agents}



\author{Dong-Ki Kim}
\affiliation{MIT-LIDS}
\affiliation{MIT-IBM Watson AI Lab}
\email{dkkim93@mit.edu}

\author{Miao Liu}
\affiliation{IBM Research}
\affiliation{MIT-IBM Watson AI Lab}
\email{miao.liu1@ibm.com}

\author{Shayegan Omidshafiei}
\affiliation{MIT-LIDS}
\affiliation{MIT-IBM Watson AI Lab}
\email{shayegan@mit.edu}

\author{Sebastian Lopez-Cot}
\affiliation{MIT-LIDS}
\affiliation{MIT-IBM Watson AI Lab}
\email{slcot@mit.edu}

\author{Matthew Riemer}
\affiliation{IBM Research}
\affiliation{MIT-IBM Watson AI Lab}
\email{mdriemer@us.ibm.com}

\author{Golnaz Habibi}
\affiliation{MIT-LIDS}
\affiliation{MIT-IBM Watson AI Lab}
\email{golnaz@mit.edu}

\author{Gerald Tesauro}
\affiliation{IBM Research}
\affiliation{MIT-IBM Watson AI Lab}
\email{gtesauro@us.ibm.com}

\author{Sami Mourad}
\affiliation{IBM Research}
\affiliation{MIT-IBM Watson AI Lab}
\email{sami.mourad@ibm.com}

\author{Murray Campbell}
\affiliation{IBM Research}
\affiliation{MIT-IBM Watson AI Lab}
\email{mcam@us.ibm.com}

\author{Jonathan P.~How}
\affiliation{MIT-LIDS}
\affiliation{MIT-IBM Watson AI Lab}
\email{jhow@mit.edu}

\renewcommand{\shortauthors}{Kim et al.}

\begin{abstract}
Collective learning can be greatly enhanced when agents effectively exchange knowledge with their peers. In particular, recent work studying agents that learn to teach other teammates has demonstrated that action advising accelerates team-wide learning. However, the prior work has simplified the learning of advising policies by using simple function approximations and only considered advising with primitive (low-level) actions, limiting the scalability of learning and teaching to complex domains. This paper introduces a novel learning-to-teach framework, called hierarchical multiagent teaching (HMAT), that improves scalability to complex environments by using the deep representation for student policies and by advising with more expressive extended action sequences over multiple levels of temporal abstraction. Our empirical evaluations demonstrate that HMAT improves team-wide learning progress in large, complex domains where previous approaches fail. HMAT also learns teaching policies that can effectively transfer knowledge to different teammates with knowledge of different tasks, even when the teammates have heterogeneous action spaces.
\end{abstract}

\keywords{Learning agent-to-agent interactions, Deep reinforcement learning}

%

\maketitle

\section{Introduction}\label{sec:introduction}
The history of human social groups provides evidence that the collective intelligence of multiagent populations may be greatly boosted if agents share their learned behaviors with others~\cite{rogers2010diffusion}. 
With this motivation in mind, we explore a new methodology for allowing agents to effectively share their knowledge and learn from other peers while maximizing their collective team reward.
Recently proposed frameworks allow for various types of knowledge transfer between agents~\cite{Taylor:2009:TLR:1577069.1755839,silva19transferSurvey}.
In this paper, we focus on transfer based on~\textit{action advising}~\cite{clouse1996integrating,torrey2013teaching,amir2016interactive,SilvaGC17,ilhan19teaching}, where an experienced ``teacher'' agent helps a less experienced ``student'' agent by suggesting which action to take next. 
Action advising allows a student to learn effectively from a teacher by directly executing suggested actions without incurring much computation overhead.

Recent works on action advising include the Learning to Coordinate and Teach Reinforcement (LeCTR) framework~\cite{omidshafiei18teach}, in which agents learn when and what actions to advise.  
While LeCTR learns peer-to-peer teaching policies that accelerate team-wide learning progress, it faces limitations in scaling to more complicated tasks with high-dimension state-action spaces, long time horizons, and delayed rewards. 
The key difficulty is \textit{teacher credit assignment}: learning teacher policies requires estimates of the impact of each piece of advice on the student agent's learning progress, but these estimates are difficult to obtain~\cite{omidshafiei18teach}. 
For example, if a student policy is represented by deep neural networks (DNN), the student requires mini-batches of experiences to stabilize its learning~\cite{Goodfellow-et-al-2016}, and these experiences can be randomly selected from a replay memory~\cite{mnih15dqn,lillicrap15ddpg}.
Hence, the student's learning progress is affected by a batch of advice suggested at varying times, and identifying the extent to which each piece of advice contributes to the student's learning is challenging (see~\Cref{sec:teacher-credit-assign} for details). 
As a result, many prior techniques use a simple function approximation, such as tile coding, to simplify the learning of advising policies~\cite{omidshafiei18teach}, but this approach does not scale well.

\noindent
\header{Contribution.}
This paper proposes a new learning-to-teach framework, called \textit{hierarchical multiagent teaching (HMAT)}.
The main contribution of this work is an algorithm that enables learning and teaching in problems with larger domains and higher complexity than possible with previous advising approaches.
Specifically, our approach: 1) handles large state-action spaces by using deep representation for student policies, 
2) provides a method based on the extended advice sequences and the use of temporary policies that correctly estimates the teacher credit assignment and improves the teacher's learning process, and 3) addresses long-horizons and delayed rewards by advising temporally extended sequences of primitive actions (i.e., sub-goal) via hierarchical reinforcement learning (HRL)~\cite{nachum18hrl}.
We note that the learning of high-level teaching policies in HMAT contrasts with prior work on action advising where teachers advise primitive (low-level) actions to students~\cite{clouse1996integrating,torrey2013teaching,amir2016interactive,SilvaGC17,omidshafiei18teach,ilhan19teaching}.
Empirical evaluations also demonstrate that agents in HMAT can learn high-level teaching policies that are transferable to different types of students and/or tasks, even when teammates have heterogeneous action spaces. 
\section{Preliminaries}
We consider a cooperative multiagent reinforcement learning (MA-RL) setting in which $n$ agents jointly interact in an environment, then receive feedback via local observations and a shared team reward.
This setting can be formalized as a partially observable Markov game, defined as a tuple $\{\mathcal{I}, \mathcal{S}, \mathcal{A}, \mathcal{T}, \mathcal{O}, \mathcal{R}, \gamma\}$ \cite{littman94markov}; 
$\mathcal{I}\!=\!\{1,\sdots,n\}$ is the set of agents, 
$\mathcal{S}$ is the set of states, 
$\mathcal{A}\!=\!\times_{i \in \mathcal{I}} \mathcal{A}^{i}$ is the set of joint actions, 
$\mathcal{T}$ is the transition probability function, 
$\mathcal{O}\!=\!\times_{i \in \mathcal{I}} \mathcal{O}^{i}$ is the set of joint observations,
$\mathcal{R}\!=\!\times_{i \in \mathcal{I}} \mathcal{R}^{i}$ is the set of joint reward functions, and 
$\gamma \in [0,1)$ is the discount factor. The cooperative setting is a specialized case of the Markov games with a shared team reward function (i.e., $\mathcal{R}^{1}=\sdots=\mathcal{R}^{n}$).
At each timestep $t$, each agent $i$ executes an action according to its policy $a_{t}^{i}\!\sim\!\pi^{i}(o^{i}_{t};\theta^i)$ parameterized by $\theta^i$, where $o^{i}_{t}\in\mathcal{O}^{i}$ is the agent $i$'s observation at timestep $t$. 
A joint action $a_{t}\!=\!\{a_{t}^{1},\sdots,a_{t}^{n}\}$ yields a transition from a current state $s_{t}\!\in\!\mathcal{S}$ to next state $s_{t+1}\!\in\!\mathcal{S}$ with probability $\mathcal{T}(s_{t+1}|s_{t},a_{t})$.
Then, a joint observation $o_{t+1}\!=\!\{o_{t+1}^{1}, \sdots, o_{t+1}^{n}\}$ is obtained and the team receives a shared reward $r_t$. 
Each agent's objective is to maximize the expected cumulative team reward $\mathbb{E}[\sum_{t}\gamma^{t}r_{t}]$.
Note that for simplicity the policy parameter will often be omitted (i.e., $\pi^{i}(o_{t}^{i};\theta^i)\!\equiv\!\pi^{i}(o_{t}^{i})$).

\subsection{Learning to Teach in Cooperative MARL}\label{sec:lectr-background}
In this section, we review key concepts and notations in the learning-to-teach framework (LeCTR)~\cite{omidshafiei18teach}.

\noindent
\header{Task-level learning problem.}
LeCTR considers a cooperative MARL setting with two agents $i$ and $j$ in a shared environment.
At each learning iteration, agents interact in the environment, collect experiences, and update their policies, $\pi^{i}$ and $\pi^{j}$, with learning algorithms, $\mathbb{L}^{i}$ and $\mathbb{L}^{j}$. 
The resulting policies aim to coordinate and optimize final task performance. 
The problem of learning task-related policies is referred to as the task-level learning problem\;\Ptask.

\noindent
\header{Advice-level learning problem.} 
Throughout the task-level learning, agents may develop unique skills from their experiences.
As such, it is potentially beneficial for agents to advise one another using their specialized knowledge to improve the final performance and accelerate the team-wide learning. 
The problem of learning teacher policies that decide \textit{when} and \textit{what} to advise is referred to as the advice-level learning problem \Padvice, where $\meta{(\cdot)}$ denotes teacher property.

\begin{figure}[t]
  \centering
  \includegraphics[trim=0 20 0 30,clip,width=\linewidth]{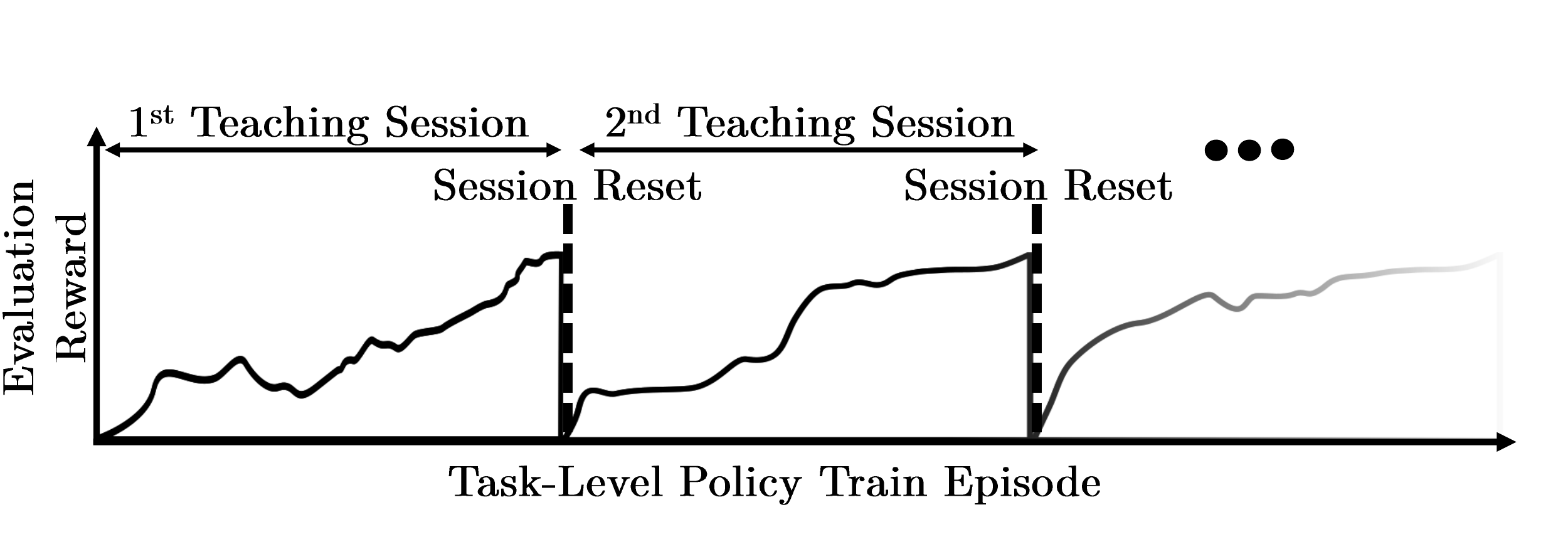}
  \caption{Illustration of task-level learning progress for each session. After completing a pre-determined episode count, task-level policies are re-initialized (teaching session reset). 
  With each new teaching session, teacher policies are better at advising students, leading to faster learning progress for task-level policies (i.e., larger area under the learning curve).}
  \label{fig:teaching-session-reward}
\end{figure}

\noindent
\header{Episode and session.} 
Learning task-level and advice-level policies are both RL problems that are interleaved within the learning-to-teach framework, but
there are important differences between \Ptask and \Padvice. 
One difference lies in the definition of learning episodes because rewards about the success of advice are naturally delayed relative to typical task-level rewards. 
For \Ptask, an episode terminates either when agents arrive at a terminal state or $t$ exceeds the maximum horizon $T$, but, for \Padvice, an episode ends when task-level policies have converged, forming one ``episode'' for learning teacher policies. 
To distinguish these two concepts, we refer to an \textit{episode} as one episode for \Ptask and a \textit{session} as one episode for \Padvice (see \cref{fig:teaching-session-reward}).

\noindent
\header{Learning complexity.} 
Another major difference between \Ptask and \Padvice is in their learning objectives.
\Ptask aims to maximize cumulative reward per episode, whereas \Padvice aims to maximize cumulative teacher reward per session, corresponding to accelerating the team-wide learning progress (i.e., a maximum area under the learning curve in a session).
Consequently, the learning complexity involved in \Padvice is significantly higher than in \Ptask due to the much longer time horizons associated with advice-level policies than task-level policies. 
For example, task-level policies need to only consider actions during one episode (i.e., the horizon of $T$ timesteps), but if the task-level policies converge after $500$ episodes, then teachers need to learn how to advise during one session (i.e., the horizon of $500\!\times\!T$ timesteps). 
In addition, teachers must consider heterogeneous knowledge (compactly represented by task-level observations, actions, and Q-values; see \Cref{sec:advice-level-policy-details}) between agents to advise, which inherently results in a higher dimension of the input size than for task-level policies and induces harder learning.

\subsection{Hierarchical Reinforcement Learning}
HRL is a structured framework with multi-level reasoning and extended temporal abstraction~\cite{feudalrl,Parr1998,sutton1999,MAXQ2000,Kulkarni16hrl,bacon16hrl,vezhnevets17hrl,abstractoptions}.
HRL efficiently decomposes a complex problem into simpler sub-problems, which offers a benefit over non-HRL in solving difficult tasks with long horizons and delayed reward assignments. 
The closest HRL framework that we use in this paper is that of~\citeauthor{nachum18hrl} \shortcite{nachum18hrl} with a two-layer hierarchical structure: the higher-level manager policy $\pi_\manager$ and the lower-level worker policy $\pi_\worker$.
The manager policy obtains an observation $o_{t}$ and plans a high-level sub-goal $g_{t} \!\sim\!\pi_\manager(o_{t})$ for the worker policy. 
The worker policy attempts to reach this sub-goal from the current state by executing a primitive action $a_{t}\!\sim\!\pi_\worker(o_{t}, g_{t})$ in the environment. 
Following this framework, an updated sub-goal is generated by the manager every $H$ timesteps and a sequence of primitive actions are executed by the worker.
The manager learns to accomplish a task by optimizing the cumulative environment reward and stores an experience $\{ o_{t}, g_{t}, \sum_{t}^{t+H-1}r_{t}, o_{t+H} \}$ every $H$ timesteps.
By contrast, the worker learns to reach the sub-goal by maximizing the cumulative intrinsic reward $r^{\text{intrinsic}}_t$ and stores an experience $\{ o_{t}, a_{t}, r^{\text{intrinsic}}_t$ $, o_{t+1} \}$ at each timestep. Without loss of generality, we also denote the next observation with the prime symbol $o^{\prime}$.
\begin{table*}[t!]
	\caption{
	Summary of considered teacher reward functions. $\hat{R}$ denotes the sum of rewards in the self-practice experiences.}
\begin{center}
    \begin{tabular}{llc}
		\toprule
		Teacher Reward Function Name & Description & Teacher Reward \\ \midrule
		Value Estimation Gain (VEG)~\cite{omidshafiei18teach}\hspace{1.5cm} & Student's $Q$-value above a threshold $\tau$ & $\mathds{1}(Q_{\text{student}}>\tau)$     \\
		Difference Rollout (DR)~\cite{xu18meta} & Difference in rollout reward before/after advising\hspace{1cm} ~~& $\hat{R}-\hat{R}_{\text{before}}$ \\
		Current Rollout (CR) & Rollout reward after advising phase & $\hat{R}$ \\ \bottomrule
	\end{tabular}
\end{center}	
\label{table:teacher-reward-table}
\end{table*}

\section{HMAT Overview}\label{sec:overview-hmat}
HMAT improves the scalability of the learning and teaching in problems with larger domains and higher complexity by employing deep student policies and learning of high-level teacher policies that decide what high-level actions to advise fellow agents and when advice should be given.
We begin by introducing our deep hierarchical task-level policy structure~in \Cref{sec:deep-HRL-task-level-policy}.
We then demonstrate why identifying which portions of the advice led to successful student learning is difficult to accomplish with deep task-level policies in~\Cref{sec:teacher-credit-assign}. 
Finally, we explain how our algorithm addresses the teacher credit assignment issue in~\Cref{sec:algorithm} and~\Cref{sec:details-of-hmat}.

\subsection{Deep Hierarchical Task-Level Policy}\label{sec:deep-HRL-task-level-policy}
\begin{figure}[t]
  \centering
  \includegraphics[width=\linewidth]{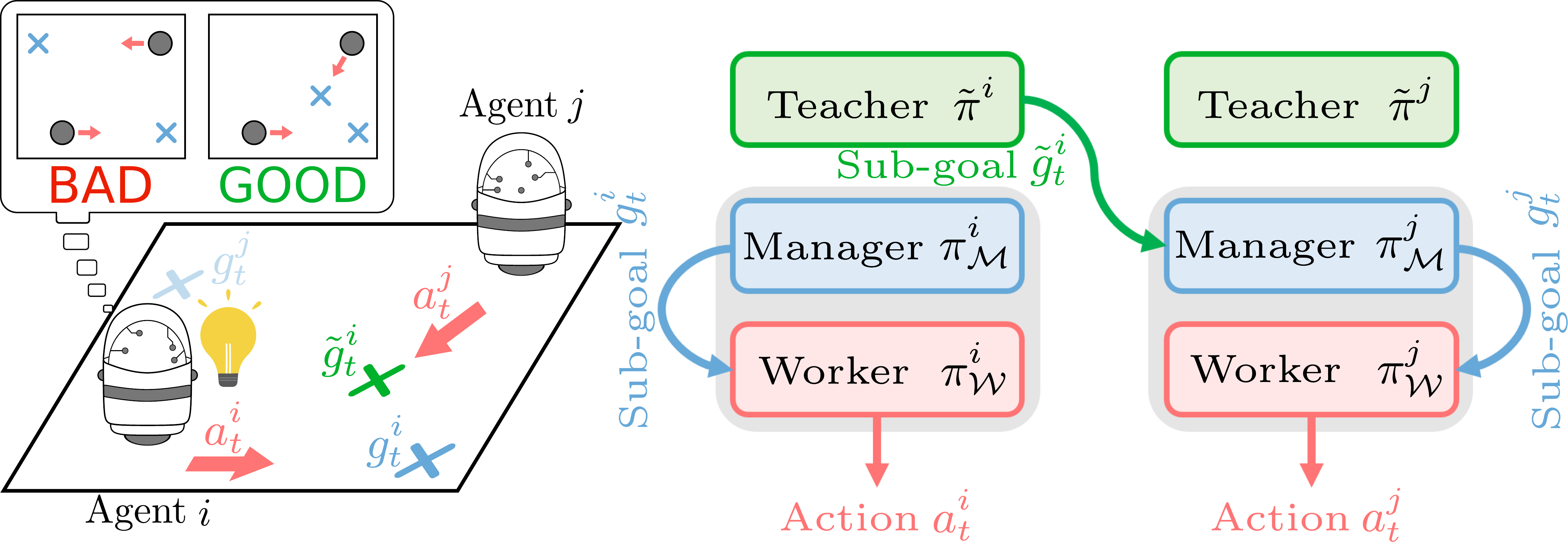}
  \caption{
  Agents teach each other according to an advising protocol (e.g., knowledgeable agent $i$ evaluates the action that agent $j$ intends to take and advises if needed).}
  \label{fig:hierarchical-teaching-overview}
\end{figure}

We extend task-level policies with DNN and hierarchical representations.
Specifically, we replace $\pi^{i}$ and $\pi^{j}$ with deep hierarchical policies consisting of manager policies, $\pi^{i}_\manager$ and $\pi^{j}_\manager$, and worker policies, $\pi^{i}_\worker$ and $\pi^{j}_\worker$ (see \cref{fig:hierarchical-teaching-overview}).
Note that the manager and worker policies are trained with different objectives.
Managers learn to accomplish a task together (i.e., solving \Ptask) by optimizing cumulative reward, while workers are trained to reach sub-goals suggested by their managers. 
In this paper, we focus on transferring knowledge at the manager-level instead of the worker-level, since manager policies represent abstract knowledge, which is more relevant to fellow agents.
Therefore, hereafter when we discuss task-level polices, we are implicitly only discussing the manager policies. 
The manager subscript $\manager$ is often omitted when discussing these task-level policies to simplify notation (i.e., $\pi^{i}\equiv\pi^{i}_{\manager}$).

\begin{figure}[t]
\captionsetup[subfigure]{skip=-1pt} 
   \begin{subfigure}[b]{0.18\textwidth}
     \centering
     \includegraphics[width=\linewidth]{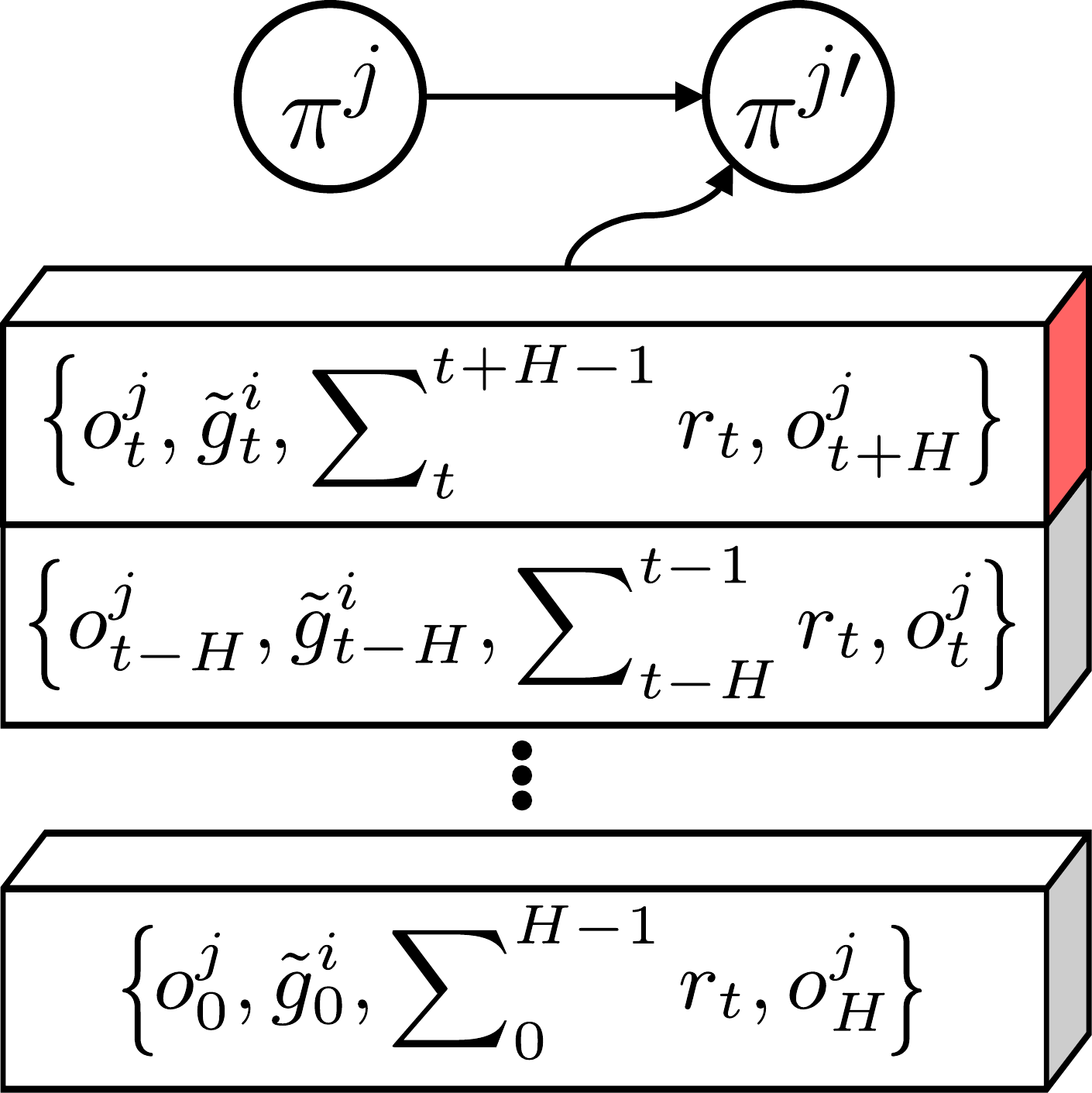}
  \caption{}
  \label{fig:credit-deep}
  \end{subfigure}
   \begin{subfigure}[b]{0.18\textwidth}
     \centering
    \includegraphics[width=\linewidth]{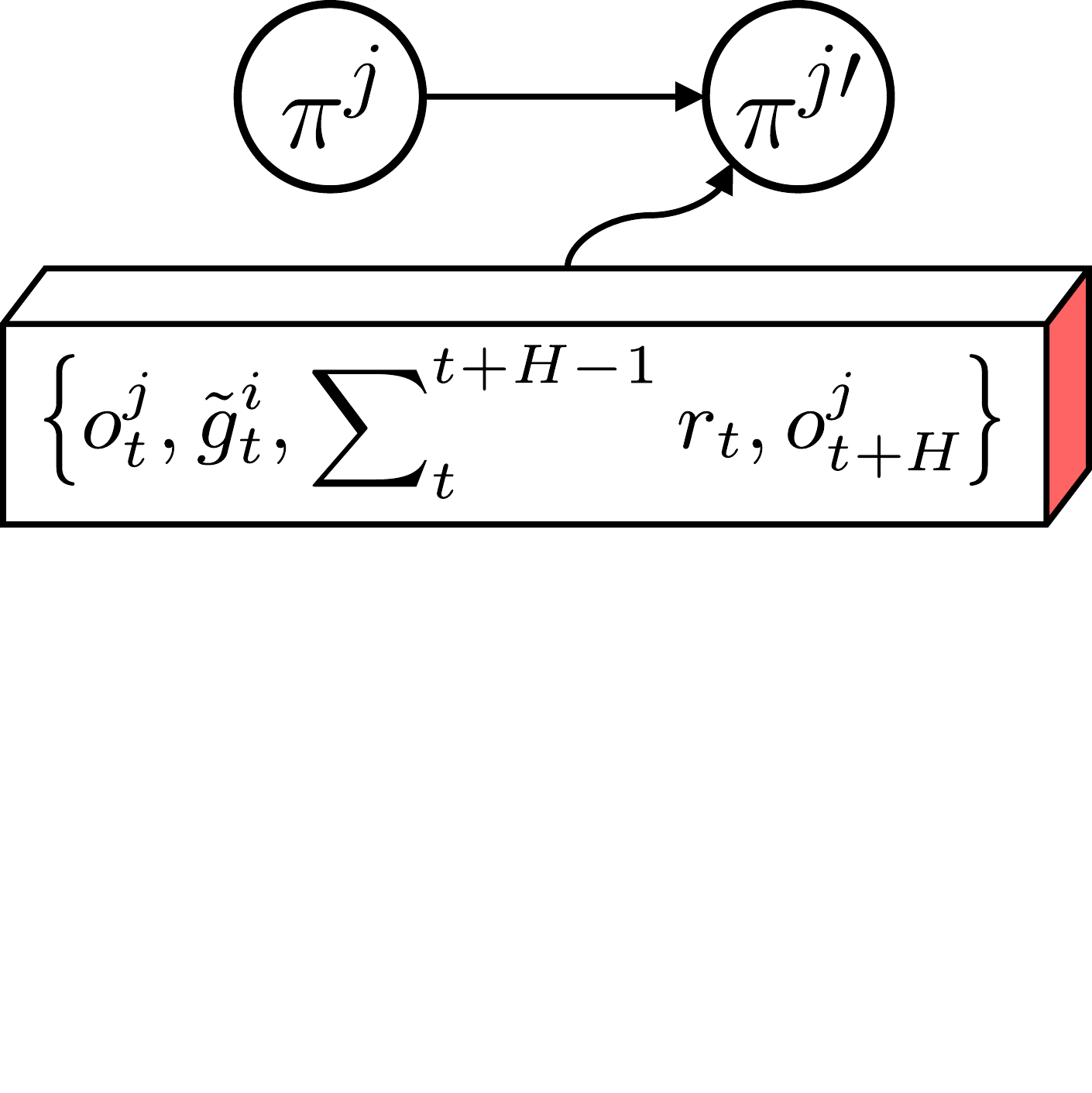}
    \caption{} 
  \label{fig:credit-tile}
	\end{subfigure}
	   \caption{
	   \textbf{(a)} A student $j$ with a deep task-level policy, which uses a mini-batch of experiences for stable training. The mini-batch includes experiences affected by both current advice (red) and previous advice (gray).
	   \textbf{(b)} A student $j$ with a tile-coding task-level policy. LeCTR assumes an online update that uses only the current experience.}
\end{figure}

\subsection{Teacher Credit Assignment Issue}\label{sec:teacher-credit-assign}
Learning advising policies requires estimates of the impact of the advice on the student agents' learning progress~\cite{omidshafiei18teach}, but these estimates are difficult to obtain especially with deep student policies.
For example, consider an example, in which a teacher agent $i$ advises a student $j$ with advice $\meta{g}^{i}_{t}$. 
Following the advice, $j$ obtains an experience of $\{o^{j}_{t}, \meta{g}^{i}_{t}, \sum_{t}^{t+H-1}r_{t}, o^{j}_{t+H}\}$.
The agent $i$ then requires feedback, reflecting the extent to which its advice $\meta{g}^{i}_{t}$ contributed to $j$'s learning progress, to be able to improve its teaching strategy. 
Note that, in general, a mini-batch of experiences is used to stably update the student's deep task-level policy from $\pi^{j}$ to $\pi^{j\prime}$~\cite{Goodfellow-et-al-2016,mnih15dqn,lillicrap15ddpg}, and this mini-batch consists of many experiences including the ones affected by both current advice $\meta{g}^{i}_{t}$ and previous advice $\{\meta{g}^{i}_{0},\sdots,\meta{g}^{i}_{t-H}\}$ (see~\Cref{fig:credit-deep}).
Consequently, $j$'s learning progress at timestep $t$ is \textit{jointly} affected by a batch of advice suggested at various times, and identifying the amount of current advice $\meta{g}^{i}_{t}$ contributes to $j$'s learning is challenging.

LeCTR simplifies the teacher credit assignment problem by assuming tile-coding task-level policies with online updates~\cite{omidshafiei18teach}. 
As such, the impact of $\meta{g}^{i}_{t}$ on $j$'s learning progress at timestep $t$ can be easily observed as the update is directly correlated by the experience affected by current advice only (see~\Cref{fig:credit-tile}). 
However, the assumption breaks with deep students due to the offline experiences, leading to incorrect credit assignment and learning of poor teachers (see~\Cref{sec:one-box-two-box-result}), which is the core challenge that our contribution addresses.

\subsection{HMAT Algorithm}\label{sec:algorithm}
HMAT iterates over the following three phases to learn how to coordinate with deep hierarchical task-level policies (i.e., solving \Ptask) and how to provide advice using the teacher policies (i.e., solving \Padvice). 
These phases are designed to address the teacher credit assignment issue with deep task-level policies (see \Cref{sec:teacher-credit-assign}). 
The issue is addressed by adapting ideas developed for learning an exploration policy for a single agent~\cite{xu18meta}, which includes an extended view of actions and the use of a temporary policy for measuring a reward for the exploration policy.
We adapt and extend these ideas from learning-to-explore in a single-agent setting into our learning-to-teach in a multiagent setting.
The pseudocode of HMAT is presented in \cref{alg:HMAT}.

\noindent
\header{Phase I: Advising.}
Agents advise each other using their teacher policies according to the advising protocol (see~\Cref{sec:advice-protocol}) during one episode.
This process generates a batch of task-level experiences influenced by the teaching policy's behavior. 
Specifically, we \textit{extend} the concept of teacher policies by providing advice in the form of a sequence of multiple sub-goals $\meta{g}_{0:T}\!=\!\{\meta{g}^{i}_{0:T}, \meta{g}^{j}_{0:T}\}$, where $\meta{g}^{i}_{0:T}\!=\!\{\meta{g}^{i}_{0}, \meta{g}^{i}_{H}, \meta{g}^{i}_{2H}, \sdots, \meta{g}^{i}_{T}\}$ denotes multiple advice by agent $i$ in one episode,
instead of just providing one piece of advice $\meta{g}_{t}\!=\!\{\meta{g}^{i}_{t},\meta{g}^{j}_{t}\}$, before updating task-level policies.
One teacher action in this extended view corresponds to providing multiple pieces of advice during one episode, which 
contrasts with previous teaching approaches that the task-level policies were updated based on a single piece of advice~\cite{clouse1996integrating,torrey2013teaching,amir2016interactive,SilvaGC17,omidshafiei18teach}.
This extension is important: by following advice $\meta{g}_{0:T}$ in an episode, a batch of task-level experiences for agents $i$ and $j$, $E^{\text{advice}}_{0:T}=\big\{\{ o^{i}_{0:T}, \meta{g}^{i}_{0:T}, r_{0:T}, o^{i\prime}_{0:T}\},\{ o^{j}_{0:T}, \meta{g}^{j}_{0:T},r_{0:T}, o^{j\prime}_{0:T} \}\big\}$, are generated to allow the stable mini-batch updates of deep policies.

\noindent
\header{Phase II: Advice evaluation.}
Learning teacher policies requires reward feedback from the advice suggested in Phase I. 
Phase II evaluates and estimates the impact of the advice on improving team-wide learning progress, yielding the teacher policies' rewards. 
We use \textit{temporary} task-level policies to estimate the teacher reward for $\meta{g}_{0:T}$, where the temporary policies denote the copy of their current task-level policies (i.e., $\pi_{\text{temp}}\!\leftarrow\!\pi$) and $\pi\!=\!\{\pi^{i},\pi^{j}\}$.
To determine the teacher reward for $\meta{g}_{0:T}$, $\pi_{\text{temp}}$ are updated for a small number of iterations using $E^{\text{advice}}_{0:T}$ (i.e., $\pi'_{\text{temp}}\!\leftarrow\!\mathbb{L}(\pi_{\text{temp}}, E^{\text{advice}}_{0:T})$), where $\mathbb{L}=\{\mathbb{L}^{i},\mathbb{L}^{j}\}$. 
Then, the updated temporary policies generate a batch of self-practice experiences $E^{\text{self-prac.}}_{0:\overline{T}}$ by rolling out a total of $\overline{T}$ timesteps without involving teacher policies.
These self-practice experiences based on $\pi'_{\text{temp}}$ reflect how agents on their own would perform after the advising phase and can be used to estimate the impact of $\meta{g}_{0:T}$ on team-wide learning.
The teacher reward function $\meta{\mathcal{R}}$ (see~\Cref{sec:teacher-reward-function}) uses the self-practice experiences to compute the teacher reward for $\meta{g}_{0:T}$ (i.e., $\meta{r}\!=\!\meta{\mathcal{R}}(E^{\text{self-prac.}}_{0:\overline{T}})$). 
The key point is that $\pi'_{\text{temp}}$, which are used to compute $\meta{r}$, are updated based on $E^{\text{advice}}_{0:T}$ only, so experiences from the past iterations do not affect the teacher reward estimation for $\meta{g}_{0:T}$.
Consequently, the use of temporary policies addresses the teacher credit assignment issue.

\noindent
\header{Phase III: Policy update.}
Task-level policies are updated to solve \Ptask by using $\mathbb{L}^{i}$ and $\mathbb{L}^{j}$, and advice-level policies are updated to solve \Padvice by using teacher learning algorithms $\meta{\mathbb{L}}^{i}$ and $\meta{\mathbb{L}}^{j}$.
In particular, task-level policies, $\pi^{i}$ and $\pi^{j}$, are updated for the next iteration by randomly sampling experiences from task-level experience memories, $\mathcal{D}^{i}$ and $\mathcal{D}^{j}$.
As in \citeauthor{xu18meta} \shortcite{xu18meta}, both $E^{\text{advice}}_{0:T}$ and $E^{\text{self-prac.}}_{0:\overline{T}}$ are added to the task-level memories.
Similarly, after adding the teacher experience, which is collected from the advising (Phase I) and advice evaluation phases (Phase II), to teacher experience memories, $\meta{\mathcal{D}}^{i}$ and $\meta{\mathcal{D}}^{j}$, teacher policies are updated by randomly selecting samples from their replay memories.
\section{HMAT Details}\label{sec:details-of-hmat}
We explain important details of the teacher policy (focusing on teacher $i$ for clarity).

\subsection{Advice-Level Policy Details}\label{sec:advice-level-policy-details}
\noindent
\header{Teacher observation and action.}
Teacher-level observations $\meta{o}_{t}=\{ \meta{o}^{i}_{t}, \meta{o}^{j}_{t} \}$ compactly provide information about the nature of the heterogeneous knowledge between the two agents.
Specifically, for agent $i$'s teacher policy, its observation $\meta{o}^{i}_{t}$ consists of:
\begin{gather*}
\meta{o}^{i}_{t}=\big\{\underbrace{o_{t}, g^{i}_{t}, g^{ij}_{t}, Q^{i}(o_{t}, g^{i}_{t}, g^{j}_{t}), Q^{i}(o_{t}, g^{i}_{t}, g^{ij}_{t})}_\text{Teacher Knowledge},\\ 
\underbrace{g^{j}_{t}, Q^{j}(o_{t}, g^{i}_{t}, g^{j}_{t}), Q^{j}(o_{t}, g^{i}_{t}, g^{ij}_{t})}_\text{Student Knowledge}, \\
\underbrace{R_{\text{phase I}}, R_{\text{phase II}}, t_{\text{remain}}}_\text{Misc.}\big\},
\end{gather*}
where $o_{t}\!=\!\{o^{i}_{t},o^{j}_{t} \}$, 
$g^{i}_{t}\!\sim\!\pi^{i}(o^{i}_{t})$, 
$g^{j}_{t}\!\sim\!\pi^{j}(o^{j}_{t})$, 
$g^{ij}_{t}\!\sim\!\pi^{i}(o^{j}_{t})$, 
$Q^{i}$ and $Q^{j}$ are the centralized critics for agents $i$ and $j$, respectively,
$R_{\text{phase I}}$ and $R_{\text{phase II}}$ are average rewards in the last few iterations in Phase I and II, respectively,
and $t_{\text{remain}}$ is the remaining time in the session.
Given $\meta{o}^{i}_{t}$, teacher $i$ decides when and what to advise, with one action for deciding whether or not it should provide advice and another action for selecting the sub-goal to give as advice.
If no advice is provided, student $j$ executes its originally intended sub-goal. 

\noindent
\header{Teacher reward function.}\label{sec:teacher-reward-function}
Teacher policies aim to maximize their cumulative teacher rewards in a session that should result in faster task-level learning progress.
Recall in Phase II that the batch of self-practice experiences reflect how agents by themselves perform after one advising phase.
Then, the question is what is an appropriate teacher reward function $\meta{\mathcal{R}}$ that can map self-practice experiences into better task-level learning performance.
Intuitively, maximizing the teacher rewards $\meta{r}$ returned by an appropriate teacher reward function means that teachers should advise so that the task-level learning performance is maximized after one advising phase.
In this work, we consider a new reward function, called current rollout (CR), which returns the sum of rewards in the self-practice experiences $E^{\text{self-prac.}}_{0:\overline{T}}$. 
We also evaluate different choices of teacher reward functions, including the ones in~\citeauthor{omidshafiei18teach} \shortcite{omidshafiei18teach} and~\citeauthor{xu18meta} \shortcite{xu18meta}, as described in~\cref{table:teacher-reward-table}.

\noindent
\header{Teacher experience.}
One teacher experience corresponds to $\meta{E}^{i}_{0:T}\!=\!\{\meta{o}^{i}_{0:T}, \meta{g}^{i}_{0:T}, \meta{r}, \meta{o}'^{i}_{0:T} \}$; where 
$\meta{o}^{i}_{0:T}=\{\meta{o}^{i}_{0},\meta{o}^{i}_{H},\meta{o}^{i}_{2H},\sdots,\meta{o}^{i}_{T}\}$ is the teacher observation;
$\meta{g}^{i}_{0:T}=\{\meta{g}^{i}_{0},\meta{g}^{i}_{H},\meta{g}^{i}_{2H},\sdots,\meta{g}^{i}_{T}\}$ is the teacher action;
$\meta{r}$ is the estimated teacher reward with $\meta{\mathcal{R}}$; and
$\meta{o}^{i\prime}_{0:T}\!=\!\{\meta{o}^{i\prime}_{0},\!\meta{o}^{i\prime}_{H},\!\meta{o}^{i\prime}_{2H},\sdots,\!\meta{o}^{i\prime}_{T}\}$ is the next teacher observation, obtained by updating $\meta{o}^{i}_{0:T}$ with the updated temporary policy $\pi^{j\prime}_{\text{temp}}$ (i.e., representing the change in student $j$'s knowledge due to advice $\meta{g}^{i}_{0:T}$).

\noindent
\header{Advice protocol.}\label{sec:advice-protocol}
Consider \cref{fig:hierarchical-teaching-overview}, where there are two roles: that of a student agent $j$ (i.e., an agent whose manager policy receives advice) and that of a teacher agent $i$ (i.e., an agent whose teacher policy gives advice). 
Note that agents $i$ and $j$ can simultaneously teach each other, but, for clarity, \cref{fig:hierarchical-teaching-overview} only shows a one-way interaction. 
Here, student $j$ has decided that it is appropriate to strive for a sub-goal $g^{j}_{t}$ by querying its manager policy. 
Before $j$ passes $g^{j}_{t}$ to its worker, $i$'s teaching policy checks $j$'s intended sub-goal and decides whether to advise or not. 
Having decided to advise, $i$ transforms its task-level knowledge into desirable sub-goal advice $\meta{g}^{i}_{t}$ via its teacher policy and suggests it to $j$. 
After $j$ accepts the advice from the teacher, the updated sub-goal $g^{j}_{t}$ is passed to $j$'s worker policy, which then generates a primitive action $a^{j}_{t}$. 

\begin{algorithm}[t!]
	\caption{HMAT Pseudocode}\label{alg:HMAT}  
	\begin{algorithmic}[1]
	    \Require Maximum number of episodes in session $S$
	    \Require Teacher update frequency $\text{f}_{\text{teacher}}$
	    \State Initialize advice-level policies $\meta{\pi}$ and 
		memories $\meta{D}$
		\For{teaching session}
		    \State Re\tikzmark{left}-initialize task-level policy parameters $\pi$
		    \State Re-initialize task-level memories $D$
		    \State Re-initialize train episode count: $e=0$
		    \While{$e \leq S$}
		        \State \tikzmark{top1}$E^{\text{advice}}_{0:T}, \{ \meta{o}_{0:T}, \meta{g}_{0:T} \}$ $\leftarrow$ Teacher's advice
		        \State Update episode count: $e \leftarrow e + 1$\tikzmark{bottom1}
		        \State \tikzmark{top2}Copy temporary task-level policies: $\pi_{\text{temp}} \leftarrow \pi$
		        \State Update to $\pi'_{\text{temp}}$ using Eqn \cref{eqn:task-critic-loss}--\cref{eqn:task-actor-loss} with $E^{\text{advice}}_{0:T}$
		        \State $E^{\text{self-prac.}}_{0:\overline{T}}$ $\leftarrow$ $\pi'_{\text{temp}}$ perform self-practice
		        \State Update episode count: $e \leftarrow e + \overline{T}/T$
		        \State \tikzmark{bottom2} $\meta{r}$ $\leftarrow$ Get teacher reward with $\meta{\mathcal{R}}$
		        \State \tikzmark{top3}Add $E^{\text{advice}}_{0:T}$ and $E^{\text{self-prac.}}_{0:\overline{T}}$ to $D$
		        \State Add a teacher experience $\meta{E}_{0:T}$ to $\meta{D}$
		        \State Update $\pi$ using Eqn \cref{eqn:task-critic-loss}--\cref{eqn:task-actor-loss} with $D$
		        \If{$e$ mod $\text{f}_{\text{teacher}}==0$}
    		        \State Update $\meta{\pi}$ using Eqn (3)--\cref{eqn:teacher-actor-loss} with $\meta{D}$
		        \EndIf\tikzmark{bottom3}
		    \EndWhile
		\EndFor
	\end{algorithmic}
	\AddNote{top1}{bottom1}{left}{Phase\\I}
	\AddNote{top2}{bottom2}{left}{Phase\\II}
	\AddNote{top3}{bottom3}{left}{Phase\\III}
    \vspace{-0.4cm}
\end{algorithm}

\subsection{Training Protocol}\label{sec:train-protocol}
\noindent
\header{Task-level training.}
We use TD3~\cite{fujimoto18td3} to train the worker and manager task-level policies. 
TD3 is an actor-critic algorithm which introduces two critics, $Q_{1}$ and $Q_{2}$, to reduce overestimation of Q-value estimates in DDPG~\cite{lillicrap15ddpg} and yields more robust learning performance. 
Originally, TD3 is a single-agent deep RL algorithm accommodating continuous spaces/actions.
Here, we extend TD3 to multiagent settings with a resulting algorithm termed MATD3, and non-stationarity in MARL is addressed by applying centralized critics/decentralized actors~\cite{foerster2017counterfactual,lowe17maddpg}. 
Another algorithm termed HMATD3 further extends MATD3 with HRL.
In HMATD3, agent $i$'s task policy critics, $Q^{i}_{1}$ and $Q^{i}_{2}$, minimize the following critic loss:
\begin{align}
\label{eqn:task-critic-loss}
\begin{split}
    \mathcal{L}&=\sum\nolimits_{\alpha=1}^{2} \E\nolimits_{\{ o,g,r,o'\}\sim \mathcal{D}^{i}}\big[y - Q_{\alpha}^{i}(o,g)\big]^{2},\\
 \text{s.t.}\quad y&=r+\gamma \min_{\beta=1,2}Q_{\beta,\text{target}}^{i}(o',\pi_{\text{target}}(o') + \epsilon),
\end{split}
\end{align}
where $o=\{o^{i},o^{j}\}$,
$g=\{g^{i},g^{j}\}$,
$o^{\prime}=\{o^{i\prime},o^{j\prime}\}$,
$\pi_{\text{target}}=\{\pi^{i}_{\text{target}}$ $, \pi^{j}_{\text{target}}\}$,
the subscript ``target'' denotes the target network,
and $\epsilon \sim \mathcal{N}(0,\sigma)$.
The agent $i$'s actor policy $\pi^{i}$ with parameter $\theta^{i}$ is updated by:
\begin{equation}
  \label{eqn:task-actor-loss}
  \nabla_{\theta^{i}}J(\theta^{i})\!=\!\E\nolimits_{o\sim \mathcal{D}^{i}}\!\!\big[\nabla_{\theta^{i}}\pi^{i}(g^{i} \vert o^{i})\!\nabla_{\!g^{i}}Q^{i}_{1}\!(o,g)\vert_{g=\pi(o)} \big].
\end{equation}

\noindent
\header{Advice-level training.}\label{sec:advice-level-training}
TD3 is also used for updating teacher policies.
We modify Eqn \cref{eqn:task-critic-loss} and \cref{eqn:task-actor-loss} to account for the teacher's \textit{extended} view.
Considering agent $i$ for clarity, agent $i$'s teacher policy critics, $\meta{Q}^{i}_{1}$ and $\meta{Q}^{i}_{2}$, minimize the following critic loss:
\begin{gather}
\begin{aligned}
\meta{\mathcal{L}}&=\text{$\sum\nolimits_{\alpha=1}^{2}$}\!\!\E\nolimits_{\meta{E}_{0:T} \sim \meta{\mathcal{D}}^{i}}\!\!\big[
 \E\nolimits_{\meta{E} \sim \meta{E}_{0:T}}\!\!\big(
 y - \meta{Q}_{\alpha}^{i}(\meta{o},\meta{g})\big)\big]^{2}, \\
  \text{s.t.}\;\;y&=\meta{r}+\gamma \min_{\beta=1,2}\meta{Q}_{\beta,\text{target}}^{i}(\meta{o}',\meta{\pi}_{\text{target}}(\meta{o}') + \epsilon),\\ \meta{E}_{0:T} &= \{ \meta{o}_{0:T},\meta{g}_{0:T},\meta{r},\meta{o}'_{0:T} \},\space\space\space
 \meta{E} = \{ \meta{o},\meta{g},\meta{r},\meta{o}' \},
\end{aligned}
\label{eqn:teacher-critic-loss}
\raisetag{35pt}
\end{gather}
where $\meta{o}=\{\meta{o}^{i},\meta{o}^{j}\}$;
$\meta{g}=\{\meta{g}^{i},\meta{g}^{j}\}$;
$\meta{o}^{\prime}=\{\meta{o}^{i\prime},\meta{o}^{j\prime}\}$;
$\meta{\pi}_{\text{target}}=\{\meta{\pi}^{i}_{\text{target}}$ $, \meta{\pi}^{j}_{\text{target}}\}$. 
The agent $i$'s actor policy $\meta{\pi}^{i}$ with parameter $\meta{\theta}^{i}$ is updated by:
\begin{equation}\label{eqn:teacher-actor-loss}
\nabla_{\meta{\theta}^{i}}J(\meta{\theta}^{i})=
\E\nolimits_{\meta{o}_{0:T}\sim \mathcal{\meta{D}}^{i}}\big[\E\nolimits_{\meta{o}\sim \meta{o}_{0:T}}\big(z\big)\big],
\end{equation}
where $z = \nabla_{\meta{\theta}^{i}}\meta{\pi}^{i}(\meta{g}^{i} \vert \meta{o}^{i})\nabla_{\meta{g}^{i}}\meta{Q}^{i}_{1}(\meta{o},\meta{g})\vert_{\meta{g}=\meta{\pi}(\meta{o})}$.
\section{Evaluation}\label{sec:experiment}
We demonstrate HMAT's performance in increasingly challenging domains that involve continuous states/actions, long horizons, and delayed rewards. 

\subsection{Experiment Setup}
Our domains are based on OpenAI's multiagent particle environment.
We modify the environment and propose new domains, called cooperative one and two box push:

\begin{figure}[t]
\captionsetup[subfigure]{skip=1pt} 
   \begin{subfigure}[b]{0.5\textwidth}
     \centering
     \includegraphics[height=0.25\linewidth]{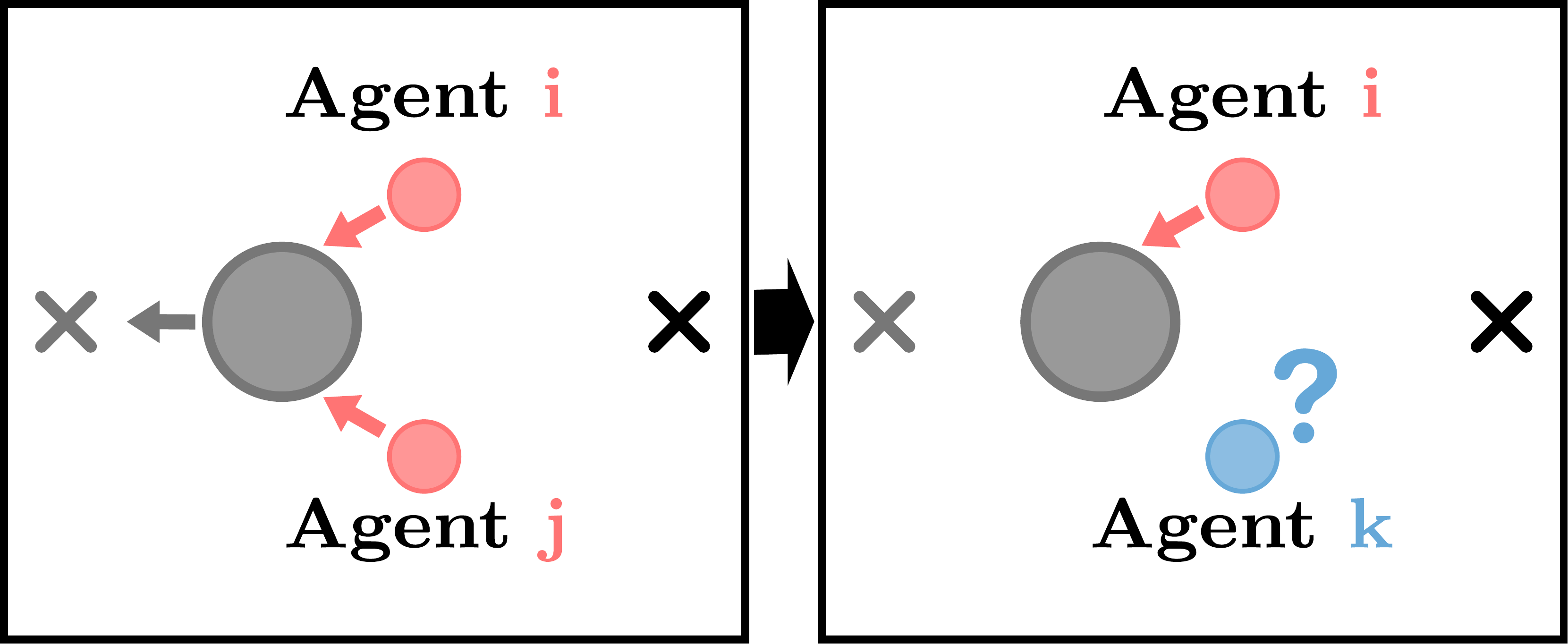}
  \caption{Cooperative one box domain.}
  \label{fig:one-box}
   \end{subfigure}
   \begin{subfigure}[b]{0.5\textwidth}
     \centering
    \includegraphics[height=0.25\linewidth]{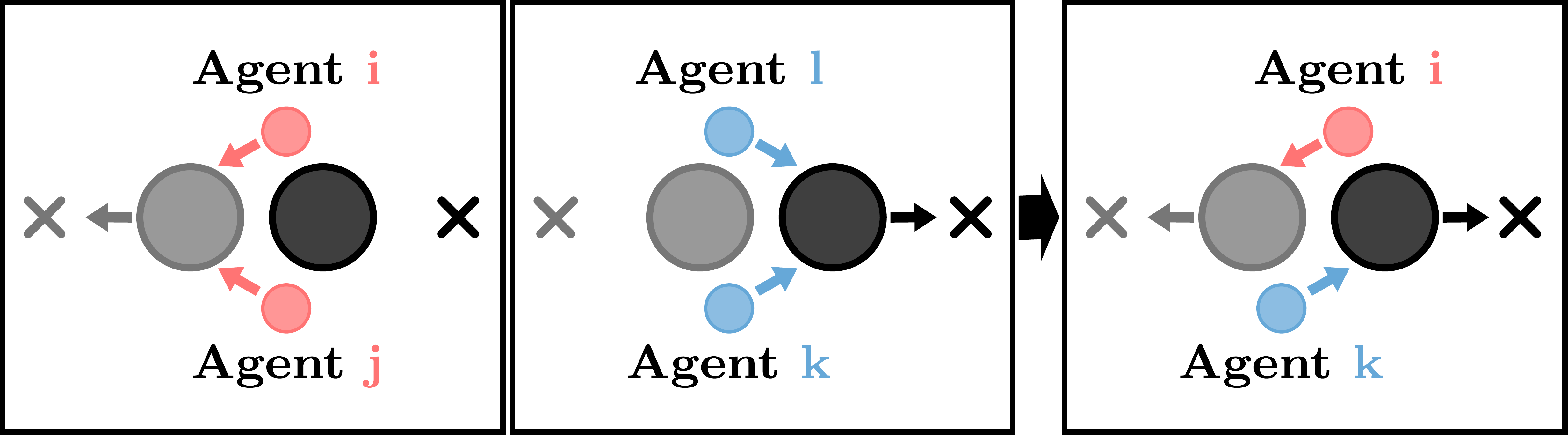}
    \put(-203, 29){{\color{white} $1$}}
    \put(-183.5, 29){{\color{white} $2$}}
    \put(-129, 29){{\color{white} $1$}}
    \put(-109, 29){{\color{white} $2$}}
    \put(-48.5, 29){{\color{white} $1$}}
    \put(-28.5, 29){{\color{white} $2$}}
  \caption{Cooperative two box domain.}
  \label{fig:two-box}
   \end{subfigure}
   \caption{Two scenarios used for evaluating HMAT.}
	\label{fig:overview_fig}
\end{figure}
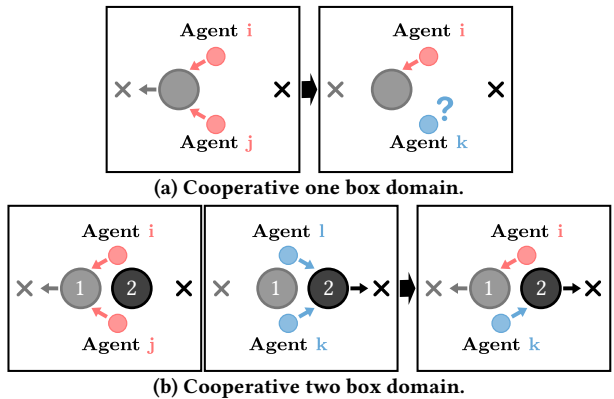

\noindent
\header{Cooperative one box push (COBP).} 
The domain consists of one round box and two agents (see \cref{fig:one-box}). 
The objective is to move the box to the target on the left side as soon as possible. 
The box can be moved iff two agents act on it together.
This unique property requires agents to coordinate.
The domain also has a delayed reward because there is no change in reward until the box is moved by the two agents. 

\noindent
\header{Cooperative two box push (CTBP).}
This domain is similar to COBP but with increased complexity. There are two round boxes in the domain (see \cref{fig:two-box}). 
The objective is to move the left box (box$1$) to the left target (target$1$) and the right box (box$2$) to the right target (target$2$).
In addition, the boxes have different mass -- box$2$ is 3x heavier than box$1$. 

\noindent
\header{Heterogeneous knowledge.} 
For each domain, we provide each agent with a different set of priors to ensure heterogeneous knowledge between them and motivate interesting teaching scenarios. 
For COBP, agents $i$ and $j$ are first trained to move the box to the left target. 
Then, agents $i$ and $k$ are teamed up, where $k$ has no knowledge about the domain. Agent $i$, which understands how to move the box to the left, should teach agent $k$ by giving good advice to improve $k$'s learning progress. 
For CTBP, agents $i$ and $j$ have received prior training about how to move box$1$ to target$1$, and agents $k$ and $l$ understand how to move box$2$ to target$2$. 
However, these two teams have different skills as the tasks involve moving boxes with different weights (light vs heavy) and also in different directions (left vs right). 
Then agents $i$ and $k$ are teamed up, and in this scenario, agent $i$ should transfer its knowledge about moving box$1$ to $k$. 
Meanwhile, agent $k$ should teach $i$ how to move box$2$, so that there is a two-way transfer of knowledge where each agent is the primary teacher at one point and primary student at another.

\noindent
\header{Domain complexity.}
COBP and CTBP are comparable to other recent MARL scenarios~\cite{wen2018probabilistic,li19minimax} but are more difficult because of the inherent delayed rewards associated with the fact that the box moves iff both agents learn to push it together. These domains are also considerably more challenging than those investigated in the prior work on learning to teach~\cite{omidshafiei18teach}, which considered domains (repeated game and gridworld) with short horizons and low-dimension, discrete space-action spaces. 

\begin{table*}[t!]
	\centering
	\small
	\caption{$\bar{V}$ and AUC for different algorithms. Results show a mean and standard deviation computed for 10 sessions. Best results in bold (computed via a t-test with p < 0.05).}
	\label{table:results_comparison}
	\renewrobustcmd{\bfseries}{\fontseries{b}\selectfont} 
	\sisetup{detect-weight,mode=text,group-minimum-digits=4}
	{
		\tabcolsep=0.15cm
		\begin{tabular}[t]{lcccccc}
			\toprule
			Algorithm & Hierarchical? & Teaching? & \multicolumn{2}{c}{One Box Push} & \multicolumn{2}{c}{Two Box Push} \\
			\cmidrule(lr){4-5} \cmidrule(lr){6-7} & & & $\bar{V}$ & AUC & $\bar{V}$ & AUC \\ \midrule
			MATD3          & \xmark & \xmark & $-12.64 \pm 3.42$ & $148 \pm 105$ & $-44.21 \pm 5.01$ & $1833 \pm 400$ \\
			LeCTR--Tile    & \xmark & \cmark & $-18.00 \pm 0.01$ & $6 \pm 0$ & $-60.50 \pm 0.05$ & $246 \pm 3\:\:$ \\
			LeCTR--D       & \xmark & \cmark & $-14.85 \pm 2.67$ & $81 \pm 80$ & $-46.07 \pm 3.00$ & $1739 \pm 223$ \\
			LeCTR--OD      & \xmark & \cmark & $-17.92 \pm 0.10$ & $17 \pm 3\:\:$ & $-60.43 \pm 0.08$ & $272 \pm 26$ \\
			AI             & \xmark & \cmark & \B{$-$11.33 $\pm$ 2.46} & $162 \pm 86\:\:$ & $-41.09 \pm 3.50$ & $2296 \pm 393$ \\
			AICI           & \xmark & \cmark & \B{$-$10.60 $\pm$ 0.85} & $200 \pm 74\:\:$ & $-38.23 \pm 0.45$ & $2742 \pm 317$ \\
			MAT (with CR)  & \xmark & \cmark & \B{$-$10.04 $\pm$ 0.38} & $274 \pm 38\:\:$ & $-39.49 \pm 2.93$ & $2608 \pm 256$ \\			
			\cmidrule(lr){1-7}
			HMATD3         & \cmark & \xmark & \B{$-$10.24 $\pm$ 0.20} & $427 \pm 18\:\:$ & $-31.55 \pm 3.51$ & $4288 \pm 335$ \\
            LeCTR--HD      & \cmark & \cmark & $-12.10 \pm 0.94$ & $265 \pm 59\:\:$ & $-38.87 \pm 2.94$ & $2820 \pm 404$ \\            
            LeCTR--OHD     & \cmark & \cmark & $-16.77 \pm 0.66$ & $70 \pm 22$ & $-60.23 \pm 0.33$ & $306 \pm 28$ \\            
            HAI            & \cmark & \cmark & \B{$-$10.23 $\pm$ 0.19} & $427 \pm 9\:\:\:\:$ & $-31.37 \pm 3.71$ & $4400 \pm 444$ \\
            HAICI          & \cmark & \cmark & \B{$-$10.25 $\pm$ 0.26} & $433 \pm 5\:\:\:\:$ & $-29.32 \pm 1.19$ & $4691 \pm 261$ \\
			HMAT (with VEG)  & \cmark & \cmark & $-$10.38 $\pm$ 0.25 & $424 \pm 28\:\:$ & $-$29.73 $\pm$ 2.89 & 4694 $\pm$ 366 \\
			HMAT (with DR)   & \cmark & \cmark & $-$10.36 $\pm$ 0.32 & $417 \pm 31\:\:$ & \B{$-$28.12 $\pm$ 1.58} & 4758 $\pm$ 286 \\
			HMAT (with CR) & \cmark & \cmark & \B{$-$10.10 $\pm$ 0.19} & \B{458 $\pm$ 8\:\:\:\:} & \B{$-$27.49 $\pm$ 0.96} & \B{5032 $\pm$ 186} \\ \bottomrule
		\end{tabular}
	}
\end{table*}
\begin{figure*}[t] \centering
\captionsetup[subfigure]{skip=-2pt} 
   \begin{subfigure}[b]{0.3\textwidth}
     \centering
     \includegraphics[width=\textwidth]{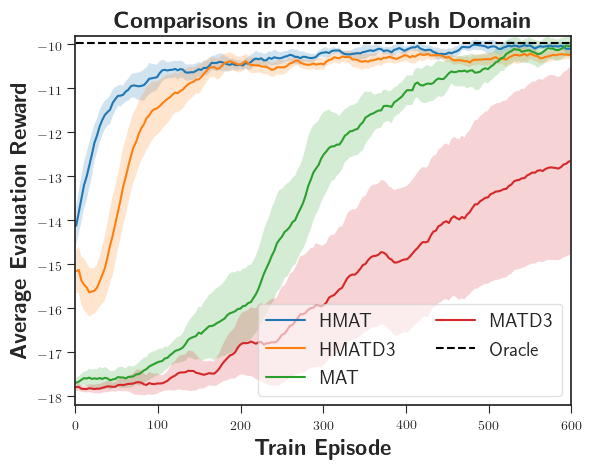}
  \caption{}
  \label{fig:one-box-result}
   \end{subfigure}
   \begin{subfigure}[b]{0.3\textwidth}
     \centering
     \includegraphics[width=\textwidth]{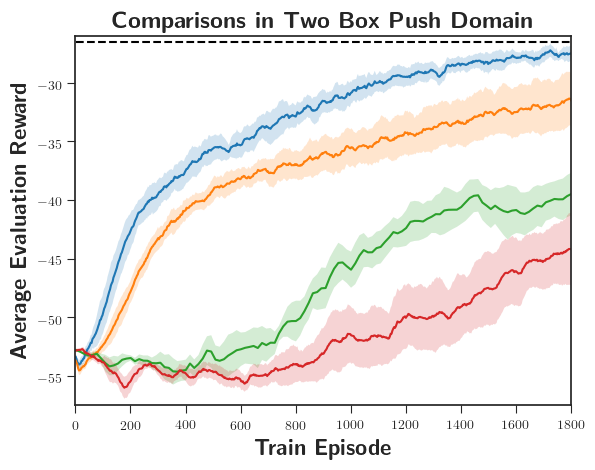}
  \caption{}
  \label{fig:two-box-result}
   \end{subfigure}
   \begin{subfigure}[b]{0.3\textwidth}
     \centering
     \includegraphics[width=\textwidth]{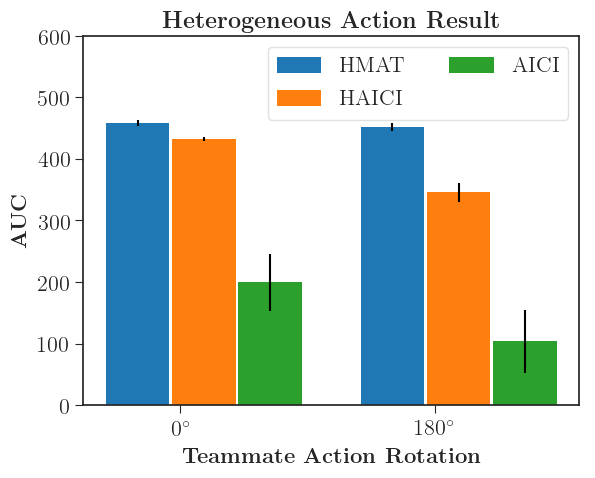}
  \caption{}
  \label{fig:hetero-action-result}
   \end{subfigure}
   \caption{
   \textbf{(a) and (b)} Task-level learning progress in COBP and CTBP, respectively. 
   The oracles in \textbf{(a)} and \textbf{(b)} refer to the performance of converged HMATD3. 
   For fair comparisons, HMAT and MAT include both the number of episodes used in Phase I and II when counting the number of train episodes. 
   \textbf{(c)} Heterogeneous action AUC based on the action rotation. Mean and 95\% confidence interval computed for 10 sessions are shown in all figures.
   }
\end{figure*}

\subsection{Baselines}\label{sec:baselines}
We compare several baselines in order to provide context for the performance of HMAT. 

\noindent
\header{No-teaching.}
MATD3 and HMATD3 (\Cref{sec:train-protocol}) are baselines for a primitive and hierarchical MARL without teaching, respectively.

\noindent
\header{Learn to teach.} 
The original LeCTR framework with tile-coding task-level policies (LeCTR--Tile) is compared.
We also consider modifications of that framework, in which task-level policies are learned with deep RL, MATD3 (LeCTR--D) and HMATD3 (LeCTR--HD).
Lastly, we additionally compare two LeCTR-based baselines using online MATD3 (LeCTR--OD) and online HMATD3 (LeCTR--OHD), where the online update denotes the task-level policy update with the most recent experience only (i.e., no mini-batch).

\noindent
\header{Heuristic teaching.} 
Two heuristic-based primitive teaching baselines, Ask Important (AI) and Ask Important--Correct Important (AICI)~\cite{amir2016interactive}, are compared.
In AI, each student asks for advice based on the importance of a state using its $Q$-values. 
When asked, the teacher agent always advises with its best action at a student state. 
Students in AICI also ask for advice, but teachers can decide whether to advise with their best action or not to advise. 
Each teacher decides based on the state importance using its $Q$-values and the difference between the student's intended action and teacher's intended action at a student state. 
Note that AICI is one of the best performing heuristic algorithms in \citeauthor{amir2016interactive} \shortcite{amir2016interactive}.
Hierarchical AI (HAI) and hierarchical AICI (HAICI) are similar to AI and AICI, but teach in the hierarchical setting (i.e., managers teach each other).

\noindent
\header{HMAT variant.} 
A non-hierarchical variant of HMAT, called MAT, is compared. 

\subsection{Results on One Box and Two Box Push}\label{sec:one-box-two-box-result}
\Cref{table:results_comparison} compares HMAT and its baselines.
The results show both final task-level performance ($\bar{V}$) (i.e., final episodic average reward measured at the end of the session) and area under the task-level learning curve (AUC) -- higher values are better for both metrics.

\noindent
\header{Comparisons to no-teaching.}
The results demonstrate improved task-level learning performance with HMAT compared to HMATD3 and with MAT compared to MATD3, as indicated by the higher final performance ($\bar{V}$) and the larger rate of learning (AUC) in \cref{fig:one-box-result,fig:two-box-result}.
These results demonstrate the main benefit of teaching that it accelerates the task-level learning progress.  

\noindent
\header{Comparisons to learning to teach.}
HMAT also achieves better performance than the LeCTR baselines. 
LeCTR--Tile shows the smallest $\bar{V}$ and AUC due to the limitations of the tile-coding representation of the policies in these complex domains. 
Teacher policies in LeCTR--D and LeCTR--HD have poor estimates of the teacher credit assignment with deep task-level policies, which result in unstable learning of the advising policies and worse performance than the no-teaching baselines (LeCTR--D vs MATD3, LeCTR--HD vs HMATD3).
In contrast, both LeCTR--OD and LeCTR--OHD have good estimates of the teacher credit assignment as the task-level policies are updated online.
However, these two approaches suffer from the instability caused by the absence of a mini-batch update for the DNN policies.

\noindent
\header{Other comparisons.}
HMAT attains the best performance in terms of $\bar{V}$ and AUC compared to the heuristics-based baselines.
Furthermore, HMAT also shows better performance than MAT, demonstrating the benefit of the high-level advising that helps address the delayed rewards and long-time horizons in these two domains.
Lastly, our empirical experiments show that CR performs the best with HMAT compared to VEG and DR.
Consistent with this observation, the learning progress estimated with CR has a high correlation with the true learning progress (see \cref{fig:teacher-reward-analysis}).
These combined results demonstrate the key advantage of HMAT in that it accelerates the learning progress for complex tasks with continuous states/actions, long horizons, and delayed rewards. 

\subsection{Transferability and Heterogeneity}\label{sec:transfer-result}
We evaluate the transferability of HMAT in advising different types of students and/or tasks and teaching with heterogeneous action spaces. 
The numerical results are mean/standard deviation for $10$ sessions and a $t$-test with $p < 0.05$ is performed for checking statistical significance.

\noindent
\header{Transfer across different student types.}
We first create a small population of students, each having different knowledge.
Specifically, we create $7$ students that can push one box to distinct areas in the one-box push domain: top-left, top, top-right, right, bottom-right, bottom, and bottom-left.
This population is divided into train $\{$top-left, top, bottom-right, bottom, and bottom-left$\}$, validation $\{$top-right$\}$, and test $\{$right$\}$ groups.
After the teacher policy has converged, we fix the policy and transfer it to a different setting in which the teacher advises a student in the test group.
Although the teacher has never interacted with the student in the test group before, it achieves an AUC of $400 \pm 10$, compared to no-teaching baseline (HMATD3) AUC of $369 \pm 27$.

\noindent
\header{Transfer across different tasks.}
We first train the teacher that learns to transfer knowledge to agent $k$ about how to move the box to the left in the one box push domain.
Then we fix the converged teacher policy and evaluate on a different task of moving the box to the right.
While task-level learning without teaching achieves AUC of $363 \pm 43$, task-level learning with teaching achieves AUC of $414 \pm 11$.
Thus, learning is faster, even when using pre-trained teacher policies from different tasks.

\noindent
\header{Teaching with heterogeneous action spaces.}
We consider heterogeneous action space variants in COBP, where agent $k$ has both remapped manager and worker action spaces (e.g., $180\degree$ rotation) compared to its teammate $i$.
Note that teachers in the heuristic-based approaches advise with their best action assuming the action space homogeneity between the teachers and students. 
Consequently, when the action space is flipped, the heuristic advising can be an inefficient teaching strategy and lead teammates to no-reward regions.
As~\cref{fig:hetero-action-result} shows, both HAICI and AICI show decreased AUC when the action space is remapped.
In contrast, the capability of learning to advise in HMAT can understand the heterogeneity in the action space, so HMAT achieves the best performance regardless of action rotation.

\subsection{HMAT Analyses}\label{hmat-analyses}
\header{Teacher reward accuracy.}
Developing ground-truth learning progress of task-level policies often requires an expert policy and could be computationally undesirable~\cite{graves17curriculum}. 
Thus, HMAT uses an estimation of the learning progress as a teacher reward.
However, it is important to understand how close the estimation is to true learning progress.
The goal of teacher policies is to maximize the cumulative teacher reward, so a wrong estimate of teaching reward would result in learning undesirable teacher behaviors. 
In this section, we aim to measure the differences between the true and estimated learning progress and analyze the CR teacher reward function, which performed the best.

\begin{figure}[t]
  \centering
  \includegraphics[height=0.6\linewidth]{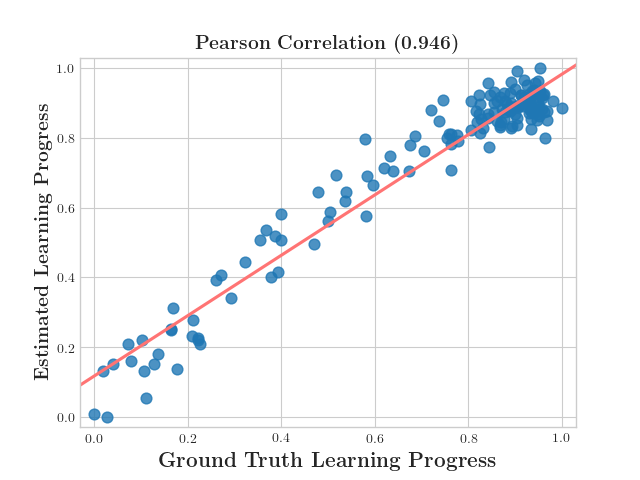}
  \caption{Ground-truth vs estimated learning progress with CR. The CR teacher reward function estimates the true learning progress with the high Pearson correlation of 0.946.}
  \label{fig:teacher-reward-analysis}
\end{figure}

In imitation learning, with an assumption of a given expert, one standard method to measure the true learning progress is by measuring the distance between an action of a learning agent and an optimal action of an expert~\cite{ross14imitation,daswani15imitation}. 
Similarly, we pre-train expert policies using HMATD3 and measure the true learning progress by the action differences.
The comparison between the true and estimated learning progress using the CR teacher reward function is shown in \cref{fig:teacher-reward-analysis}. 
The Pearson correlation is $0.946$, which empirically shows that CR well estimates the learning progress. 

\begin{figure}[t]
  \centering
  \includegraphics[height=0.6\linewidth]{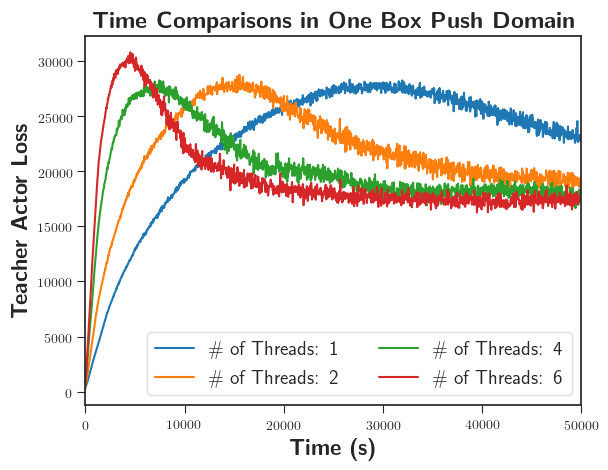}
  \caption{
  Teacher actor loss between the different numbers of threads. 
  With an increasing number of threads, a teacher policy converges faster.
  }
  \label{fig:multi-thread-hmat}
\end{figure}

\noindent
\header{Asynchronous HMAT.}
Similar to a deep RL algorithm requiring millions of episodes to learn a useful policy~\cite{mnih15dqn}, our teacher policies would require many \textit{sessions} to learn. 
As one session consists of many episodes, much time might be needed until teacher policies converge.
We address this potential issue with an asynchronous policy update with multi-threading as in asynchronous advantage actor-critic (A3C)~\cite{mniha16A3C}.
A3C demonstrated a reduction in training time that is roughly linear in the number of threads.
We also show that our HMAT variant, asynchronous HMAT, achieves a roughly linear reduction in training time as a function of the number of threads (see \cref{fig:multi-thread-hmat}).
\section{Related Work}
Action advising is not the only possible approach to transfer knowledge.
Works on imitation learning studies on how to learn a policy from expert demonstrations~\cite{ross11imitation,ross14imitation,daswani15imitation}.
Recent work applied imitation learning for multiagent coordination~\cite{le2017coordinated} and explored effective combinations of imitation learning and HRL \cite{le2018hierarchical}.
Curriculum learning~\cite{bengio2009curriculum,tsvetkov16,graves17curriculum}, which progressively increases task difficulty, is also relevant.
Approaches in curriculum learning measure or learn the hardness of tasks and design a curriculum for a learning agent to follow.
Curriculum learning has many applications, including recent works that learn a training data curriculum for an image classifier~\cite{fan2018learning,jiang18c}.
Another work of \citeauthor{thomaz06hri} \shortcite{thomaz06hri} studies transferring human knowledge to an RL agent via providing the reward signals.
While most works on these topics focus on learning and/or transferring knowledge for solving single-agent problems, this paper investigates peer-to-peer knowledge transfer in cooperative MARL.
The related work by \citeauthor{xu18meta} \shortcite{xu18meta} learns an exploration policy, which relates to our approach of learning teaching policies.
However, their approach led to unstable learning in our setting, motivating our policy update rule in Eqn
\cref{eqn:teacher-critic-loss} and \cref{eqn:teacher-actor-loss}.
We also consider various teacher reward functions, including the one in \citeauthor{xu18meta} \shortcite{xu18meta}, and a new reward function of CR that is empirically shown to perform better in our domains.

\section{Conclusion}
The paper presents HMAT, which utilizes the deep representation and HRL to transfer knowledge between agents in cooperative MARL. 
We propose a method to overcome the teacher credit assignment issue and show accelerated learning progress in challenging domains.
In the future, we will extend this work along with the following directions: 1) investigating ways to expand HMAT and associated learning to teach methods to problems involving more than two agents, 2) studying theoretical properties of HMAT, and 3) applying HMAT to more complex domains that may require more than two levels of hierarchies.

\section*{Acknowledgements}
Research funded by IBM (as part of the MIT-IBM Watson AI Lab initiative) and computational support through Amazon Web Services. Dong-Ki Kim was also supported by a Kwanjeong Educational Foundation Fellowship.

\appendix
\section{Appendix}
\subsection{Implementation Details}
Each policy's actor and critic are two-layer feed-forward neural networks consisting of the rectified linear unit (ReLU) activations.
In terms of the task-level actor policies, a final layer of the tanh activation, which 
has the range of (-1, 1), is used at the output.
The actor policies for both the worker and primitive task-level policies output two actions that correspond to x--y forces to move in the one box and two box domains.
Similarly, the actor policies for the manager task-level policies output two actions, but they correspond to sub-goals of (x, y) coordinate.
Regarding the advice-level actor policies, they output four actions, where the first two outputs correspond to \textit{what} to advise (i.e., continuous sub-goal advice of (x, y) coordinate) and the last two outputs correspond to \textit{when} to advise (i.e., discrete actions converted to the one-hot encoding).
Two separate final layers of the tanh activation and the linear activation are applied to what and when to advise actions, respectively.
The Gumbel-Softmax estimator~\cite{jang2016categorical} is used to compute gradients for the discrete actions of when to advise. 
Regarding the hierarchical methods, since we focus teaching at the manager-level, not at the worker-level, we pre-train the worker policies by giving randomly generated sub-goals and then fix the policies.
The intrinsic reward function to pre-train the worker policy is the negative distance between the current position and sub-goal.
All hierarchical methods presented in this work use the pre-trained workers.

\subsection{COBP/CTBP Experiment Details}
Each agent's observation includes its position/speed, the positions of the box(es), targets, and its peer. 
In terms of initialization, the left and right box initialize at ($-0.25,0.0$) and ($0.25,0.0$), respectively, the two targets initialize at ($-0.85,0.0$) and ($0.85,0.0$), respectively, and agents reset at random locations. Regarding the maximum timestep $T$, COBP and CTBP have $50$ and $100$ timesteps, respectively. 
The maximum number of episodes in a session $S$ is $600$ episodes for COBP and $1800$ episodes for CTBP. 
Also, for the self-practice rollout size $\overline{T}$, COBP and CTBP use $100$ and $200$ timesteps, respectively.
Managers in all hierarchical methods generate the sub-goal every $H=5$ timesteps. 
astly, we use Adam optimizer with the actor learning rate of $0.0001$ and the critic learning rate of $0.001$.

\bibliographystyle{ACM-Reference-Format}  
\bibliography{references}  

\end{document}